%% file: main.tex
\documentclass[sigconf]{acmart}

\usepackage{fancyhdr}

\usepackage{tikz}
\usepackage[pages=some]{background}
\usepackage[utf8]{inputenc} % allow utf-8 input
\usepackage[T1]{fontenc}    % use 8-bit T1 fonts
\usepackage{booktabs}       % professional-quality tables
\usepackage{amsfonts}       % blackboard math symbols
\usepackage{nicefrac}       % compact symbols for 1/2, etc.
\usepackage{xcolor}         % colors
\usepackage{color}
\usepackage{ulem}

\usepackage{url}

\usepackage{booktabs} % for professional tables

\usepackage{multirow}
\usepackage{subcaption}

\usepackage{enumitem}

\usepackage{wrapfig}

\usepackage{collcell}
\usepackage{xstring}
\usepackage{datatool}
\usepackage{booktabs}
\usepackage{environ}

\usepackage{color}
\definecolor{darkblue}{rgb}{0.0, 0.2, 0.6}

\usepackage{balance}
\usepackage{float}
\usepackage{listings}
\usepackage{color}
\definecolor{dkgreen}{rgb}{0,0.6,0}
\definecolor{gray}{rgb}{0.5,0.5,0.5}
\definecolor{mauve}{rgb}{0.58,0,0.82}
\definecolor{bg}{rgb}{0.9,0.9,0.9}

\usepackage[capitalize,noabbrev]{cleveref}

\theoremstyle{plain}

\theoremstyle{definition}

\theoremstyle{remark}

\usepackage{cite}
\usepackage{amsmath,amsfonts}
\usepackage{algorithmic}
\usepackage{graphicx}
\usepackage{textcomp}
\usepackage{xcolor}
\def\BibTeX{{\rm B\kern-.05em{\sc i\kern-.025em b}\kern-.08em
    T\kern-.1667em\lower.7ex\hbox{E}\kern-.125emX}}
\usepackage{algorithm}
\usepackage{algorithmic}

\usepackage{booktabs} % for professional tables
\usepackage{multirow} 

\usepackage{bbm}

\usepackage{twoopt}
\usepackage{tikz}
\usetikzlibrary{fit}

\newcommandtwoopt\Textbox[5][2.5cm][2cm]{%
\begin{tikzpicture}[remember picture,overlay]
  \coordinate (aux) at ([xshift=#1]#4);
  \node[inner ysep=3pt,yshift=0.6ex,draw=green,thick,
    fit=(#3) (aux),baseline] 
    (box) {};
  \node[text width=#2,anchor=north east,
    font=\sffamily\footnotesize,align=right] 
    at (box.north east) {#5};
\end{tikzpicture}%
}
\DeclareCaptionLabelFormat{alglabel}{Alg.~#2}
\usepackage{xspace}

\newcommand{\harp}{\textsc{harp}\xspace}
\newcommand{\gnndse}{\textsc{gnn-dse}\xspace}
\newcommand{\pmodel}{\textsc{compareXplore}\xspace}
\newcommand{\ndt}{\textsc{Node Difference Attention}\xspace}

\newcommand{\diffone}{\textsc{$\mathrm{d_1}$}\xspace}
\newcommand{\difftwo}{\textsc{$\mathrm{d_2}$}\xspace}
\newcommand{\diffthree}{\textsc{$\mathrm{d_3}$}\xspace}
\newcommand{\diffall}{\textsc{$\mathrm{ALL}$}\xspace}

\newcommand{\hlsyn}{\textsc{HLSyn}\xspace}
\newcommand{\numDSEKernels}{six\xspace}

\usepackage[colorinlistoftodos]{todonotes}

\definecolor{applegreen}{rgb}{0.55, 0.71, 0.0}

\AtBeginDocument{%
  \providecommand\BibTeX{{%
    \normalfont B\kern-0.5em{\scshape i\kern-0.25em b}\kern-0.8em\TeX}}}

\setcopyright{acmlicensed}

\copyrightyear{2024}
\acmYear{2024}
\setcopyright{rightsretained}
\acmConference[MLCAD '24]{2024 ACM/IEEE International Symposium on Machine Learning for CAD}{September 9--11, 2024}{Salt Lake City, UT, USA}
\acmBooktitle{2024 ACM/IEEE International Symposium on Machine Learning for CAD (MLCAD '24), September 9--11, 2024, Salt Lake City, UT, USA}
\acmDOI{10.1145/3670474.3685940}

\acmISBN{979-8-4007-0699-8/24/09}

\begin{document}

\title{Learning to Compare Hardware Designs for High-Level Synthesis}

\author{Yunsheng Bai\textsuperscript{1,2},
Atefeh Sohrabizadeh\textsuperscript{2},
Zijian Ding\textsuperscript{2},
Rongjian Liang\textsuperscript{1},
Weikai Li\textsuperscript{2}, 
Ding Wang\textsuperscript{2},
Haoxing Ren\textsuperscript{1},
Yizhou Sun\textsuperscript{2},
Jason Cong\textsuperscript{2}}

\affiliation{
  \institution{\textsuperscript{1}NVIDIA Corporation, \textsuperscript{2}University of California, Los Angeles}
  \country{United States of America}
}

\email{yunshengb@nvidia.com, {atefehsz,bradyd}@cs.ucla.edu, rliang@nvidia.com, weikaili@cs.ucla.edu}
\email{allenwang2333@gmail.com, haoxingr@nvidia.com, {yzsun,cong}@cs.ucla.edu}

\input{sec-abstract.tex}




\received{29 July 2024}
\received[revised]{29 July 2024}
\received[accepted]{14 Aug 2024}

\maketitle
% \backgroundsetup{opacity=1, scale=1, angle=0, contents={
% \begin{tikzpicture}[remember picture, overlay]
% \node[anchor=north east, inner xsep=50pt, inner ysep=10pt] at (current page.north east) {
% \href{https://www.acm.org/publications/policies/artifact-review-and-badging-current}{
% \includegraphics[width=50pt]{vis/artifacts_available_v1.1.pdf}
% }};
% \end{tikzpicture}
% }}
% \BgThispage

\pagestyle{fancy}
\fancyhf{}

% Even pages
\fancyhead[RE]{\scriptsize Yunsheng Bai et al.}
\fancyhead[LE]{\scriptsize MLCAD '24, September 9--11, 2024, Salt Lake City, UT, USA}

% Odd pages
\fancyhead[LO]{\scriptsize Learning to Compare Hardware Designs for High-Level Synthesis}
\fancyhead[RO]{\scriptsize MLCAD '24, September 9--11, 2024, Salt Lake City, UT, USA}

% \begin{IEEEkeywords}
% High-Level Synthesis, Design Space Exploration, Graph Neural Networks, Electronic Design Automation
% \end{IEEEkeywords}

\input{sec-intro.tex}
\input{sec-related.tex}

\input{sec-model.tex}

\input{sec-result.tex}
\input{sec-conc.tex}

\begin{acks}
This work was partially supported by NSF grants 2211557, 1937599, 2119643, and 2303037, SRC JUMP 2.0 PRISM Center, NASA, Okawa Foundation, Amazon Research, Cisco, Picsart, Snapchat, and the CDSC industrial partners (https://cdsc.ucla.edu/partners/). The authors would also like to thank Robert Kirby (Nvidia) for initial discussions and model training.
\end{acks}

\bibliographystyle{ACM-Reference-Format}
\bibliography{bibliography}

\end{document}

%% file: sec-abstract.tex
\begin{abstract}

High-level synthesis (HLS) is an automated design process that transforms high-level code into optimized hardware designs, enabling the rapid development of efficient hardware accelerators for various applications such as image processing, machine learning, and signal processing. To achieve optimal performance, HLS tools rely on pragmas, which are directives inserted into the source code to guide the synthesis process, and these pragmas can have various settings and values that significantly impact the resulting hardware design. State-of-the-art ML-based HLS methods, such as \harp, first train a deep learning model, typically based on graph neural networks (GNNs) applied to graph-based representations of the source code and its pragmas. They then perform design space exploration (DSE) to explore the pragma design space, rank candidate designs using the trained model, and return the top designs as the final designs. However, traditional DSE methods face challenges due to the highly nonlinear relationship between pragma settings and performance metrics, along with complex interactions between pragmas that affect performance in non-obvious ways. 

To address these challenges, we propose \pmodel, a novel approach that learns to compare hardware designs for effective HLS optimization. \pmodel introduces a hybrid loss function that combines pairwise preference learning with pointwise performance prediction, enabling the model to capture both relative preferences and absolute performance values. Moreover, we introduce a novel \ndt module that focuses on the most informative differences between designs, enhancing the model's ability to identify critical pragmas impacting performance. \pmodel adopts a two-stage DSE approach, where a pointwise prediction model is used for the initial design pruning, followed by a pairwise comparison stage for precise performance verification. Experimental results demonstrate that \pmodel achieves significant improvements in ranking metrics and generates high-quality HLS results for the selected designs, outperforming the existing state-of-the-art method.

\end{abstract}

%% file: sec-intro.tex
\section{Introduction} %\YS{the introduction is too long. compress the DSA part.}
\label{sec-intro}

High-Level Synthesis (HLS) has emerged as a transformative technology in the realm of hardware design, bridging the gap between high-level software abstractions and efficient hardware implementations. Tools such as Xilinx's Vitis HLS~\citep{vitis_hls} automate the translation of high-level code into optimized hardware designs, enabling rapid development of specialized accelerators for image processing, machine learning, signal processing~\citep{cong2011high,canis2011legup,nane2015survey,cong2022fpga}, etc.

Central to the HLS design flow is the concept of pragmas—directives embedded within the high-level code that guide the synthesis process, which heavily affects the effectiveness of HLS in producing high-quality designs. 
% The pragma design space is vast and complex, with a multitude of possible settings and combinations that can dramatically impact the resulting hardware latency, resource utilization, and power consumption.
However, the relationship between pragma settings and performance metrics is highly nonlinear, with intricate interactions and dependencies that are difficult to predict or reason about. 
Traditional design space exploration (DSE) methods, which rely on heuristics and iterative synthesis, often fall short in efficiently identifying optimal configurations~\citep{zuo2013improving}.

To address these challenges, researchers have turned to machine learning (ML) techniques to aid in the DSE process. State-of-the-art ML-based HLS methods, such as \gnndse~\citep{sohrabizadeh2022automated} and \harp~\citep{sohrabizadeh2023robust}, utilize deep learning models to guide the DSE process for high-quality pragma configurations. These approaches typically involve two key steps: (1) training a predictive model, often based on graph neural networks (GNNs)~\citep{kipf2016semi,wu2020comprehensive}, to learn the mapping between designs with varying pragma settings and performance metrics, and (2) performing DSE using the trained model to rank and select the most promising candidate designs.

While ML-based methods have shown promise in improving the efficiency and effectiveness of HLS, they still face limitations in capturing the complex relationships and interactions within the pragma design space. We hypothesize that the highly nonlinear nature of the design space, coupled with the intricate dependencies between pragmas, poses challenges for accurate performance prediction and the ranking of candidate designs. 
% {\as{this part needs a supporting example and further discussion. first to show that previous models have this problem and then to show that our model doesn't and why.}}  
Our experiments in Section~\ref{sec-result} provide further evidence of this issue.

We hypothesize that comparing a pair of designs and predicting which design is better may an easier task compared with accurately predicting the design quality. In this paper, we propose \pmodel, a novel approach to DSE in HLS that leverages the power of comparative learning, a paradigm where models are trained to discern relative preferences between data points~\citep{burges2005learning}, to navigate the pragma design space effectively. \pmodel introduces a hybrid loss function that combines pairwise preference learning with pointwise performance prediction, enabling the model to capture both relative preferences and absolute performance values. By learning to compare designs based on their pragma settings and performance characteristics, \pmodel can effectively identify the most promising configurations and guide DSE towards optimal hardware designs.

Moreover, we introduce a novel \ndt module that focuses on the most informative differences between designs. This attention mechanism allows \pmodel to prioritize the pragma settings that have the greatest impact on performance, enhancing the model's ability to make accurate comparisons and identify critical design choices.

To balance exploration and exploitation in DSE, \pmodel uses a two-stage approach. In the first stage, a pointwise prediction model is used to explore and efficiently prune the design space, identifying a subset of promising candidates. This stage leverages the model's ability to estimate absolute performance values and quickly eliminate suboptimal designs. In the second stage, a pairwise comparison model is leveraged to perform precise performance verification and rank the remaining candidates based on their relative performance.
% \as{is it accurate to call this stage exploration?}. 
This stage takes advantage of the model's comparative learning capabilities to make nuanced 
% \as{what does fine-grained mean here? maybe use another word since it has a very special meaning in hardware community} 
distinctions between designs and select the top-performing configurations.

We evaluate \pmodel on a comprehensive set of HLS kernels and demonstrate its effectiveness in improving the quality of the generated hardware designs. Experimental results show that \pmodel achieves significant improvements in ranking metrics, compared to existing state-of-the-art methods. Moreover, the designs selected by \pmodel consistently outperform those obtained through baseline approaches.

The main contributions of this paper are as follows: We propose \pmodel, a two-stage approach to DSE in HLS 
% \as{the DSE part is actually not novel, we are using the same algorithm as before. we just directly use a ranking metric instead of ranking the designs by their latency value} 
that leverages comparative learning to effectively navigate the pragma design space and identify high-quality hardware designs.\footnote{Our code is available at \url{https://github.com/NVlabs/CompareXplore}.}
\begin{itemize}
    \item We introduce a hybrid loss function that combines pairwise preference learning with pointwise performance prediction, enabling \pmodel to capture both relative preferences and absolute performance values.
    \item We present a novel \ndt module that focuses on the most informative differences between designs, enhancing the model's ability to identify critical pragma settings and make accurate comparisons.
    \item We propose a two-stage DSE approach that balances exploration and exploitation, using a pointwise prediction model for efficient design pruning and a pairwise comparison model for precise performance verification inspired by Ranked Choice Voting (RCV)~\citep{fairvote_rcv,pildes2021legality}.
    \item We conduct extensive experiments on a diverse set of HLS benchmarks and demonstrate the superiority of \pmodel over \harp in terms of ranking metrics and the quality of the generated hardware designs.
\end{itemize}

% The remainder of this paper is organized as follows. Section 2 provides an overview of related work in HLS, DSE, and ML-based approaches. Section 3 describes the proposed \pmodel framework, including the hybrid loss function, Node Difference Attention module, and two-stage DSE approach. Section 4 presents the experimental setup, benchmark details, and evaluation metrics. Section 5 discusses the experimental results and analysis. Finally, Section 6 concludes the paper and outlines future research directions.

%% file: sec-related.tex
\section{Related Work} %\YS{the introduction is too long. compress the DSA part.}
\label{sec-related}

% \subsection{Machine Learning for EDA}

% Machine learning (ML) algorithms are gaining popularity in the EDA domain due to their ability to efficiently solve complex, often NP-complete problems and produce high-quality solutions~\citep{huang2021machine,ren2022machine,ren2022embracing}. ML and deep learning (DL) models have shown success in various EDA stages, including high-level synthesis (HLS)~\citep{ustun2020accurate, sohrabizadeh2023robust, bai2023towards}, logic synthesis~\citep{neto2019lsoracle}, physical design~\citep{lu2022advancing}, verification and signoff~\citep{zhang2020grannite}, automated code generation~\citep{liu2023chipnemo,liu2023verilogeval,zhong2023llm4eda,chang2024data}, analog layout~\citep{hakhamaneshi2022pretraining}, etc.

% \subsection{GNN for EDA}

% Graph Neural Networks (GNNs) have proven particularly effective in EDA tasks~\citep{ma2022towards, zeng2019graph}. Their ability to learn representations for graph-structured data makes them well-suited for representing programs, Boolean functions, netlists, and layouts commonly encountered in EDA problems~\citep{ma2022towards, zeng2019graph}. GNNs have been applied in HLS to improve accuracy in performance prediction~\citep{bai2023harp, ustun2020predicting} and mapping operations to FPGA resources~\citep{ustun2020predicting}.

\subsection{ML and GNN for HLS and DSE}

Traditional ML algorithms like random forests and linear regression have been used to model HLS tools~\citep{meng2016adaptive}. However, recent studies demonstrate that GNNs significantly improve accuracy in various HLS tasks~\citep{sohrabizadeh2022automated, wu2021ironman, wu2022ironman, wu2022high}. 
% For instance, GNN-DSE~\citep{bai2021gnn} presents a graph representation of programs with pragmas and uses a GNN-based model to predict latency and resource utilization. Other works like~\citep{ustun2020predicting, li2021ironman, wu2022hignn} focus on predicting resource mapping or critical paths using GNNs on DFGs.

% \subsection{HLS Design Space Exploration (DSE)}

Learning algorithms have been applied to accelerate HLS DSE for finding Pareto-optimal designs~\citep{wu2021ironman}. In contrast to traditional heuristics-based approaches~\citep{sohrabizadeh2022autodse}, these methods employ data-driven techniques. IronMan~\citep{wu2021ironman}, for instance, trains a reinforcement learning agent to optimize resource allocation.

\subsection{Pairwise Comparison in ML}

Pairwise comparison has a broad range of applications in machine learning beyond its traditional uses in ranking items~\citep{negahban2012iterative,wauthier2013efficient,shah2018simple}. In fields such as information retrieval~\citep{liu2009learning} and recommender systems~\citep{rokach2012initial}, pairwise methods have proven effective for sorting and prioritizing items based on user preferences~\citep{burges2005learning}. 
% For instance, the RankNet model by \citep{burges2005learning} leverages pairwise comparisons to learn item rankings, which has been influential in developing subsequent ranking models.

% Pairwise comparison and learning to rank (LTR) are fundamental paradigms within machine learning, with wide-ranging applications in information retrieval, recommender systems, and other domains.  The core idea of pairwise comparison is to learn a model that can assess the relative preference or ranking between two items.

% \textbf{Ranking} \enspace In web search, ranking algorithms such as RankNet~\citep{burges2005learning}  learn to order documents based on relevance to a query. Similarly, recommender systems use pairwise preferences from user interactions to rank items by predicted user satisfaction~\citep{liu2009learning}.

\textbf{Metric Learning} \enspace In metric learning, pairwise comparisons are used to learn meaningful distances between items. Techniques such as contrastive loss and triplet loss are used to learn a distance metric in which similar items are closer~\citep{hoffer2015deep, masana2018metric}.
% , supporting tasks such as face recognition and anomaly detection~\citep{hoffer2015deep, masana2018metric}

% This is often achieved by using pairwise comparisons to directly optimize the embedding space.

\begin{figure*}[h]
\centering
\includegraphics[width=0.95\textwidth]{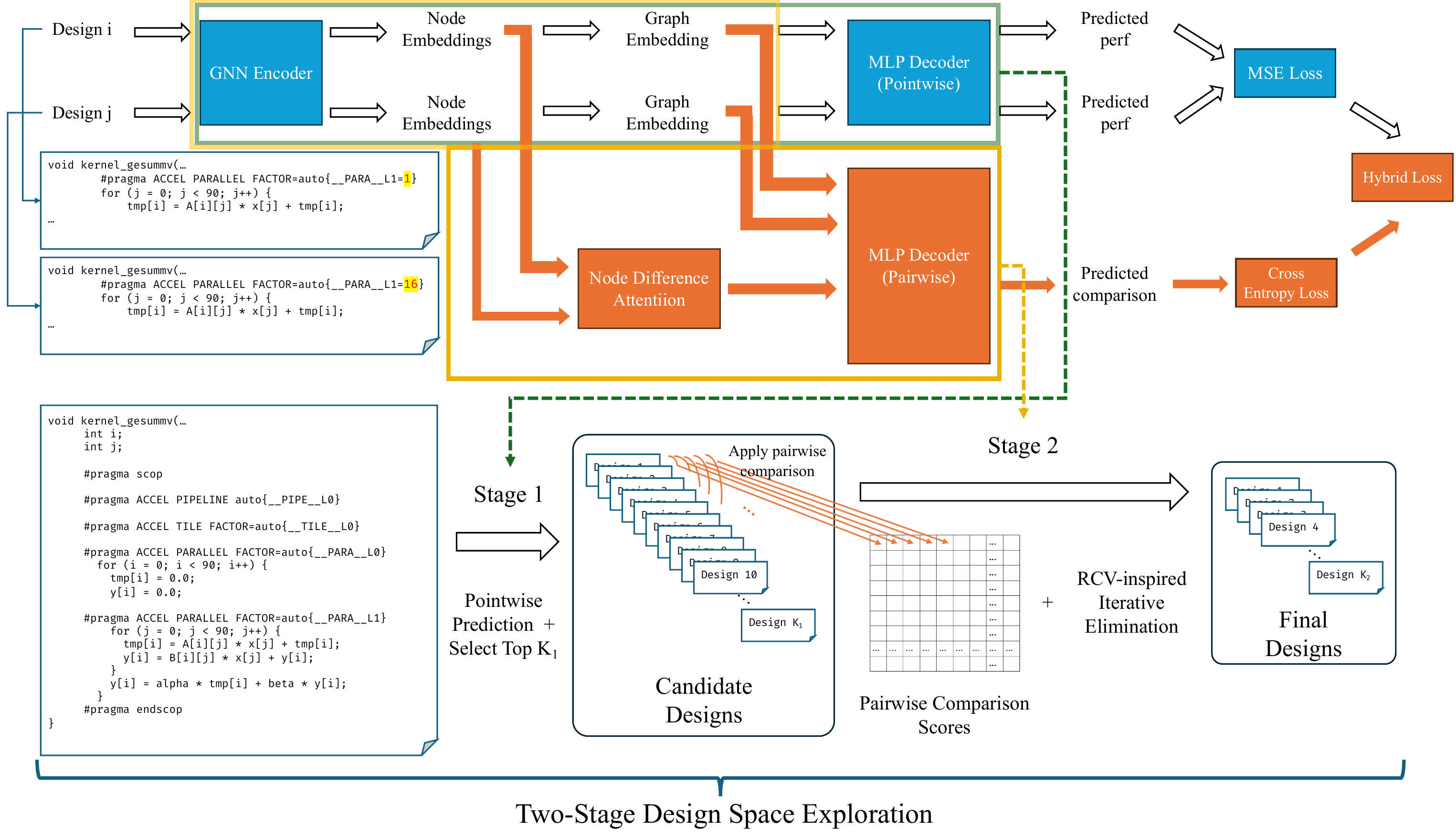}
\vspace{-0.3cm}
\caption{Overview of \pmodel. The model consists of a GNN encoder, a \ndt module, and two MLP decoders for pairwise comparison and pointwise prediction tasks. The GNN encoder learns node embeddings by aggregating information from neighboring nodes. The \ndt module focuses on the most informative differences between node embeddings, computes attention scores based on these differences, and aggregates the embedding differences. The model is used in the two-stage DSE process depicted at the bottom.
% The pairwise MLP decoder takes the concatenated graph-level embeddings and their differences as input to predict the relative performance of two designs. The pointwise MLP decoder takes individual design embeddings as input to predict the absolute performance metric for each design. 
% During training, a hybrid loss function combining pairwise preference learning and pointwise performance prediction is used to capture both relative preferences and absolute performance values. 
The major novel components are highlighted in the reddish color.}
\label{fig:model_architecture}
\end{figure*}

\textbf{Preference Learning} \enspace Pairwise comparison is also central to preference learning~\citep{furnkranz2010preference}, where the goal is to learn a model that predicts preferences between items based on observed pairwise comparisons.
% , which is useful in voting systems or selection processes.

\textbf{Natural Language Processing.} \enspace Pairwise comparison methods are crucial for tasks such as sentence similarity~\citep{sun2022sentence,wang2023going}, evaluating machine translation~\citep{guzman2019pairwise}, and aligning Large Language Models with human preferences~\citep{ouyang2022training,rafailov2024direct,song2024preference}.

% While less common in HLS, learning to compare has been explored in hardware optimization. Hartnett et al.~\citep{hartnett2024learning} demonstrate a machine learning based method that ranks logically equivalent quantum circuits based on their expected performance, leading to significant improvements in reducing noise and increasing fidelity. This demonstrates the potential of learning-to-rank techniques beyond traditional applications, and inspires their adaptation to the problem of optimizing design configurations in HLS. 

% While not common in HLS, pairwise comparison is established in ML for tasks like ranking, metric learning, and recommender systems~\citep{burges2005learning,  hoffer2015deep, furnkranz2010preference}. RankNet~\citep{burges2005learning}, a neural network model, is widely used for learning to rank items based on pairwise preferences.

\subsection{Pairwise Comparison in Electronic Design Automation}

In the context of hardware optimization, comparative learning and ranking approaches have been explored in the quantum computing domain for optimizing circuit layouts~\citep{hartnett2024learning}, which demonstrates a machine learning based method that ranks logically equivalent quantum circuits based on their expected performance, leading to improvements in reducing noise and increasing fidelity. To the best of our knowledge, we are among the first to adopt the pairwise comparison paradigm in ML-based HLS.

% This highlights the potential of learning-to-rank techniques beyond traditional applications, and its relevance to optimizing design configurations in HLS.

%% file: sec-model.tex
\section{Methodology} 
\label{sec-model}

\subsection{Overview}
In this section, we introduce our proposed model, \pmodel, for effective design space exploration in high-level synthesis. \pmodel is a novel comparative learning framework that combines pointwise prediction and pairwise comparison models to efficiently navigate the pragma design space and identify high-quality design configurations.

\subsection{Problem Setup}

Let $\mathcal{G} = (\mathcal{V}, \mathcal{E})$ denote the graph representation of an HLS design used by \harp~\citep{sohrabizadeh2023robust}, where $\mathcal{V}$ is the set of nodes and $\mathcal{E}$ is the set of edges. Each node $v \in \mathcal{V}$ is associated with a feature vector $\mathbf{x}_v \in \mathbb{R}^d$, where $d$ is the feature dimension. We denote $|\mathcal{G}|$ as the number of nodes.

The goal of design space exploration (DSE) is to find the optimal valid pragma configuration $\mathbf{p}^* \in \mathcal{P}$ that maximizes the performance metric $y$, which is intuitively the inverse of the latency defined consistently with \citep{sohrabizadeh2023robust}, where $\mathcal{P}$ is the space of all possible pragma configurations.

\subsection{Model Architecture}
\pmodel consists of a GNN encoder and an MLP-based decoder, as shown in Figure~\ref{fig:model_architecture}.

\textbf{GNN Encoder:} The GNN encoder learns node embeddings by aggregating information from neighboring nodes. We adopt a stack of GNN layers, such as GCN~\citep{kipf2016semi}, GAT~\citep{velickovic2017graph}, GIN~\citep{xu2018powerful}, or TransformerConv~\citep{shi2020masked}, to capture the structural information and node features of the hardware design graph. The output of the GNN encoder is a set of node embeddings ${\mathbf{h}_v \in \mathbb{R}^{d'} | v \in \mathcal{V}}$, where $d'$ is the embedding dimension. A pooling operation such as summation is applied to the node embeddings to obtain one embedding per design denoted as $\mathbf{h}_{\mathcal{G}}$.

\textbf{Node Difference Attention (\ndt):} The \ndt module is designed to focus on the most informative differences between node embeddings. It takes the node embeddings from the GNN encoder and computes attention scores based on the differences between node pairs. The attention scores are then used to weight the differences, emphasizing the most critical pragma-related differences. The weighted differences are aggregated to obtain a graph-level embedding. 

For a design pair $(i,j)$, denote their node embeddings as $\mathbf{H}_i \in \mathbb{R}^{|\mathcal{G}_i| \times d'}$ and $\mathbf{H}_j \in \mathbb{R}^{|\mathcal{G}_j| \times d'}$.
% During training, we sample a mini-batch of $n$ designs where $n$ is even, denoted as $\mathcal{B} = \{\mathcal{G}_1, \mathcal{G}_2, \ldots, \mathcal{G}_n\}$, where each $\mathcal{G}_i$ represents a design graph. To efficiently compare designs within the mini-batch, we split the node embeddings into two halves:
% \begin{equation}
% \mathbf{H}_1 = {\mathbf{h}_1, \mathbf{h}_2, \ldots, \mathbf{h}_{n/2}},\quad
% \mathbf{H}_2 = {\mathbf{h}_{n/2+1}, \mathbf{h}_{n/2+2}, \ldots, \mathbf{h}_n}
% \end{equation}
% where $\mathbf{h}_i \in \mathbb{R}^{d'}$ is the embeddings of design $i$. This splitting allows us to compare design $\mathcal{G}_i$ with design $\mathcal{G}_{i+n/2}$ for $i \in {1, 2, \ldots, n/2}$, simplifying the implementation and reducing computational overhead.
Since DSE is only concerned with designs of the same kernel, during training, we only compare designs of the same kernel, i.e. $\mathcal{G}_i$ and $\mathcal{G}_j$ only differ in the pragma nodes and $|\mathcal{G}_i| = |\mathcal{G}_j|$.
The differences between the node embeddings in $\mathbf{H}_i$ and $\mathbf{H}_j$ are computed: $\mathbf{D}_{ij} = \mathbf{H}_i - \mathbf{H}_j$.
The attention scores are computed by concatenating the node embeddings with their corresponding differences and passing them through an attention network:
\begin{equation}
\mathbf{s}_{ij} = \text{AttentionNet} \big(\text{concat}(\mathbf{H}_i, \mathbf{H}_j, \mathbf{D}_{ij})\big),\quad
\mathbf{s}_{ij} \in \mathbb{R}^{|\mathcal{G}_i|}.
\end{equation}
$\text{AttentionNet}$ is a multi-layer perceptron (MLP) that produces attention scores.
To focus on the most informative differences, we propose the following attention mechanism to learn which node-level embedding difference contributes the most to the comparison between the designs:
\begin{equation}
\mathbf{a}_{ij} = \text{softmax}(\mathbf{s}_{ij}),\quad
\mathbf{a}_{ij} \in \mathbb{R}^{|\mathcal{G}_i|}.
\end{equation}
The attention scores are then used to weight the difference embeddings and aggregate the differences:
\begin{equation}
\mathbf{h}_{\mathcal{G}_{ij}} = \sum_{k=1}^{|\mathcal{G}_i|} \text{softmax}(\mathbf{s}_{ij})_{k} \cdot \mathbf{D}_{ij,k} 
\end{equation}
where $|\mathcal{G}_i| = |\mathcal{G}_j|$ as described previously, and $k$ indicates the $k$-th element in a $\text{softmax}(\mathbf{s}_{ij})$ and the $k$-th row in $\mathbf{D}_{ij}$. $\mathbf{h}_{\mathcal{G}_{ij}} \in \mathbb{R}^{d'}$ can be viewed as the graph-level difference-embedding that captures the most informative pragma-related differences.

\textbf{MLP Decoders:} Since there are two tasks, the pairwise design comparison task and the pointwise design prediction task, we use two MLP decoders.

For the pairwise prediction task, the MLP decoder takes the concatenated results of various comparison operations applied to the graph-level embeddings of the two designs as input. Given the graph-level difference-embeddings $\mathbf{h}_{\mathcal{G}_{ij}}$ produced by the \ndt module, and the individual graph-level embeddings $\mathbf{h}_{\mathcal{G}_{i}}$ and $\mathbf{h}_{\mathcal{G}_{j}}$ produced by the pooling operation over $\mathbf{H}_i$ and $\mathbf{H}_j$:
\begin{equation}
\mathbf{h}_{\text{pair}_{ij}} = \text{concat}(\mathbf{h}_{\mathcal{G}_{i}} \odot \mathbf{h}_{\mathcal{G}_{j}},
\mathbf{h}_{\mathcal{G}_{ij}}),
\end{equation}
where $\odot$ denotes the Hadamard product, i.e. element-wise product.

For the pointwise prediction task, the MLP decoder takes the individual design embedding $\mathbf{h}_{\mathcal{G}_{i}}$ and $\mathbf{h}_{\mathcal{G}_{j}}$ as input:
\begin{align}
\mathbf{h}_{\text{point}_{i}} = \mathbf{h}_{\mathcal{G}_{i}}, \\
\mathbf{h}_{\text{point}_{j}} = \mathbf{h}_{\mathcal{G}_{j}}.
\end{align}

Both MLP decoders consist of multiple fully connected layers with non-linear activations, such as ReLU. The pairwise MLP decoder outputs a 2-dimensional vector representing the raw logits for the pairwise comparison, while the pointwise MLP decoder outputs a scalar value representing the predicted performance metric for the individual design, i.e.
\begin{align}
\mathbf{z}_{ij} = \text{MLP}_{pair}(\mathbf{h}_{\text{pair}_{ij}}), \\
\mathbf{z}_{i} = \text{MLP}_{point}(\mathbf{h}_{\text{point}_{i}}), \\
\mathbf{z}_{j} = \text{MLP}_{point}(\mathbf{h}_{\text{point}_{j}}).
\end{align}

\subsection{Training of \pmodel with Hybrid Loss Function}

To train \pmodel, we propose a hybrid loss function that combines pairwise preference learning with pointwise performance prediction. This enables the model to capture both relative preferences between designs and absolute performance values.

The hybrid loss function is defined as:

\begin{equation}
\mathcal{L} =  \mathcal{L}_{point} + \alpha \mathcal{L}_{pair}
\end{equation}

where $\alpha \in [0, 1]$ is a hyperparameter that controls the balance between the pairwise and pointwise losses.

The pairwise loss $\mathcal{L}_{pair}$ is calculated using a cross-entropy loss that evaluates the model's ability to correctly rank pairs of design configurations. For each pair of designs, the MLP decoder outputs a 2-D vector of logits, $\mathbf{z}_{ij}$, indicating the model's confidence in the relative performance of designs $i$ and $j$. The softmax function is applied to these logits to obtain probabilities, $\mathbf{p}_{ij}=[\log(p_{ij}^{(1)}),\log(p_{ij}^{(2)})]$.

The cross-entropy loss for the pairwise comparison is defined as:
\begin{align}
\mathcal{L}_{pair} &= -\sum_{(i,j) \in \mathcal{D}} \left( \mathbbm{1}(y_i > y_j) \log(p_{ij}^{(1)}) \right. \nonumber \\
&\quad \left. + \mathbbm{1}(y_i \leq y_j) \log(p_{ij}^{(2)}) \right),
\end{align}
where $\mathbbm{1}(\cdot)$ is the indicator function, and $\mathcal{D}$ is the set of all pairs $(i,j)$ sampled during training. 

The pointwise loss $\mathcal{L}_{point}$ is computed using a mean squared error (MSE) loss between the ground-truth and the predicted performance metric over design pairs sampled for $\mathcal{L}_{pair}$.

\subsection{Two-Stage DSE Approach}
We propose two-stage approach for design space exploration, as described in Algorithm~\ref{alg:two_stage_dse}.

\begin{algorithm}[H]
\caption{Two-Stage Design Space Exploration}
\label{alg:two_stage_dse}
\begin{algorithmic}[1]
\STATE \textbf{Stage 1:} Use the pointwise prediction model to score and prune the design space to the top $K_1$ candidate $designs$ as done in \citep{sohrabizadeh2023robust}.
\STATE \textbf{Stage 2:} Apply pairwise comparisons among the pruned $designs$ to obtain $K_2 < K_1$ final designs.

% \STATE $n \gets K_1$
\STATE $scores \gets \text{Initialize array of size } K_1 \times K_1 \text{ with zeros}$
\FOR{$i \gets 1$ to $K_1$}
\FOR{$j \gets i+1$ to $K_1$}
\STATE $d_i \gets designs[i]$
\STATE $d_j \gets designs[j]$
\STATE $scores[d_i][d_j] \gets \text{PairwiseComparisonModel}(d_i, d_j)$

\ENDFOR
\ENDFOR

\STATE $remain \gets designs$
\WHILE{$|remain| > K_2$}
\STATE $points \gets \text{Initialize array of size } |remain| \text{ with zeros}$
\FOR{$i \gets 1$ to $|remain|$}
\STATE $d_i \gets remain[i]$
\FOR{$j \gets i+1$ to $|remain|$}
% \IF{$i \neq j$}
\STATE $d_j \gets remain[j]$
\STATE $points[d_i] \gets points[d_i] + scores[d_i][d_j]$
\STATE $points[d_j] \gets points[d_j] + 1 - scores[d_i][d_j]$
% \ENDIF
\ENDFOR
\ENDFOR
\STATE $minPoints \gets \min(points)$
\STATE $remain \gets remain \setminus minPoints$
\ENDWHILE

\RETURN $remain$
\end{algorithmic}
\end{algorithm}

\begin{table*}[h]
\centering
\caption{Main results. The numbers in parentheses indicate the number of designs used for regression (525) and the number of design pairs used for classification (174 for $\diffone$, 827 for $\difftwo$, 1411 for $\diffthree$, and 8139 for $\diffall$).}
\label{tab:results}
\begin{tabular}{lcccccccc}
\toprule
\multirow{2}{*}{Model} & \multicolumn{1}{c}{Regression ($\downarrow$)} & \multicolumn{4}{c}{Classification ($\uparrow$)} & \multicolumn{1}{c}{Ranking ($\uparrow$)} \\
\cmidrule(lr){2-2} \cmidrule(lr){3-6} \cmidrule(lr){7-7}
& RMSE (525) & ACC, $\diffone$ (174) & ACC, $\difftwo$ (827) & ACC, $\diffthree$ (1411) & ACC, $\diffall$ (8139) & Kendall's $\tau$  \\
\midrule
\harp & \textbf{0.2333} & \textbf{0.8218} & 0.8609 & 0.8866 & 0.8859 & 0.4157 \\
\pmodel & 0.3570 & 0.8161 & \textbf{0.8888} & \textbf{0.9118} & \textbf{0.9117} & \textbf{0.4319} \\
\bottomrule
\end{tabular}
\end{table*}

In the first stage, the pointwise prediction model is used to efficiently prune the design space $\mathcal{P}$ and identify a subset of $K_1$ promising candidate designs $\mathcal{P}' \subset \mathcal{P}$. This stage leverages the model's ability to estimate absolute performance values and quickly eliminate suboptimal designs.

In the second stage, the pairwise comparison model is used to perform precise performance verification on the candidate designs in $\mathcal{P}'$. The pairwise comparisons are used to rank the designs based on their relative performance, and the top-performing designs are selected as the final designs.

Specifically, the design with the minimum total score is iteratively removed from the $remain$ set until only $K_2$ designs are left. This process is analogous to the Ranked Choice Voting (RCV) system, where the candidate with the fewest votes is eliminated in each round. In our case, the pairwise comparison scores serve as the "votes" that determine which designs are eliminated and which ones remain in the top $K_2$ set. This approach allows for a more nuanced decision-making process.

\subsection{Complexity Analysis}

Compared to the original \harp model or a pointwise-only approach, the newly introduced \ndt module has a time complexity of $\mathcal{O}(|\mathcal{G}|d')$, which is linear to the number of nodes.

In the two-stage DSE approach, the pointwise prediction stage has the same time complexity as \harp. The newly introduced pairwise comparison stage has a time complexity of $\mathcal{O}(K_1^2)$, where $K_1$ is the number of candidate designs which is usually set to 100. The actual pairwise comparison, i.e. lines 4-10 in Algorithm~\ref{alg:two_stage_dse}, can be performed in a batch-wise fashion, with a batch size of $B$. This reduces the time complexity to $\mathcal{O}(K_1^2/B)$.

%% file: sec-result.tex
\section{Experiments} %\YS{the introduction is too long. compress the DSA part.}
\label{sec-result}

\subsection{Experimental Setup}
We evaluate our proposed approach using the Xilinx Vitis HLS tool (version 2021.1), which is a more recent version compared to the one used in \harp and the \hlsyn benchmark~\citep{bai2023towards}. Our dataset consists of 40 kernels from various application domains. The kernels are synthesized using Vitis HLS, and the resulting designs are used for training and evaluation.

The dataset consists of a total of 10,868 designs, with 9,818 (90.34\%) used for training, 525 (4.83\%) for validation, and 525 (4.83\%) for testing. The test set is used as a transductive test set, where the model has access to the design graphs but not their performance values during training. We ensure all the sampled design pairs come from the training set for a fair comparison. The validation loss is used to select the best model for testing. Training is conducted on a machine with 8 NVIDIA PG506 GPUs.

\subsection{Hyperparameter and Implementation Details}

Our approach adopts TransformerConv with 7 layers and 64-dimensional node embeddings. Consistent with \harp, we use node attention and encode pragmas using MLPs. The model is trained using the AdamW optimizer with a learning rate of 0.001 and a batch size of 128 for 1600 epochs. We use a cosine learning rate scheduler~\citep{loshchilov2016sgdr}. The prediction target, performance, is defined consistently with \harp. $\alpha$ is set to 1. For DSE, we set $K_1$ to 100 and $K_2$ to 10 with a total time budget of 12 hours with a batch size $B=512$. The model is implemented in PyTorch Geometric~\citep{Fey/Lenssen/2019}. The full hyperparameters, model implementation, and datasets will be released publicly to enhance reproducibility. 

\subsection{Evaluation Metrics}
We evaluate our approach using two main metrics: pairwise classification accuracy and ranking metrics.
\begin{itemize}
    \item Pointwise Regression Error: This metric measures the model's ability to accurately make prediction for the performance metric for each design in the test set. The Root Mean Squared Error (RMSE) is used.
    \item Pairwise Classification Accuracy: This metric measures the model's ability to correctly predict which design in a pair has better performance across design pairs in the test set. We report the accuracy for different degrees of pragma differences (\diffone, \difftwo, \diffthree indicating design pairs differing by 1, 2 and 3 pragma settings) and the overall accuracy (\diffall).
    \item Ranking metric: We report Kendall's $\tau$, which measures the ordinal association between the predicted performance rankings and the true performance rankings of the designs in the test\citep{kendall1938new}. 
\end{itemize}

We randomly select \numDSEKernels kernels where we perform DSE followed by running the HLS tool to evaluate the selected designs by the DSE process. We report the lowest latency of the selected $K_2$ designs.

\subsection{Loss Curves}
\begin{figure}[h]
    \centering
    \includegraphics[width=0.7\columnwidth]{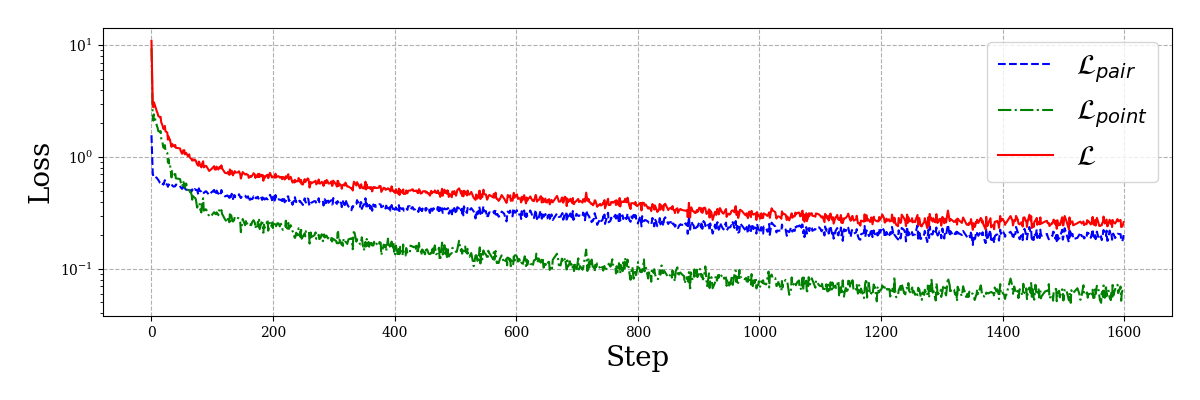}
    \vspace{-0.5cm}
    \caption{Loss curves for the proposed \pmodel. }
    \label{fig:loss}
\end{figure}

Figure~\ref{fig:loss} presents the training loss curves for our proposed \pmodel, including the pairwise loss ($\mathcal{L}_{pair}$), pointwise loss ($\mathcal{L}_{point}$), and the overall loss ($\mathcal{L}$). The figure demonstrates a decrease in both pairwise and pointwise losses, indicating the model's effectiveness in learning from the data for both tasks.

\subsection{Main Results}
Table~\ref{tab:results} presents the comparison between the vanilla \harp model and our proposed approach. Our approach achieves higher pairwise classification accuracy for designs with more pragma differences (\difftwo and \diffthree) and overall (\diffall) compared to the vanilla \harp model. In terms of ranking performance, our approach achieves a higher Kendall's $\tau$ score, indicating a better alignment between the predicted and true rankings of the designs. While \harp achieves a lower regression error, the hybrid loss design in \pmodel leads to a more balanced performance across classification accuracy and ranking metrics.

The accuracy improves as the number of differences between designs increases (
\diffone $<$ \difftwo $<$ \diffthree). This suggests that the model finds it easier to distinguish between designs that have more differences. When designs differ in more pragmas, the performance metrics tend to vary more significantly, making it easier for the model to learn and identify which design is better. The increasing accuracy from \diffone to \diffthree suggests potential future work, such as incorporating curriculum learning to progressively improve the model's performance on more challenging design pairs with smaller performance differences~\citep{bengio2009curriculum}.

\begin{figure}[h]
    \centering
    \includegraphics[width=0.55\columnwidth]{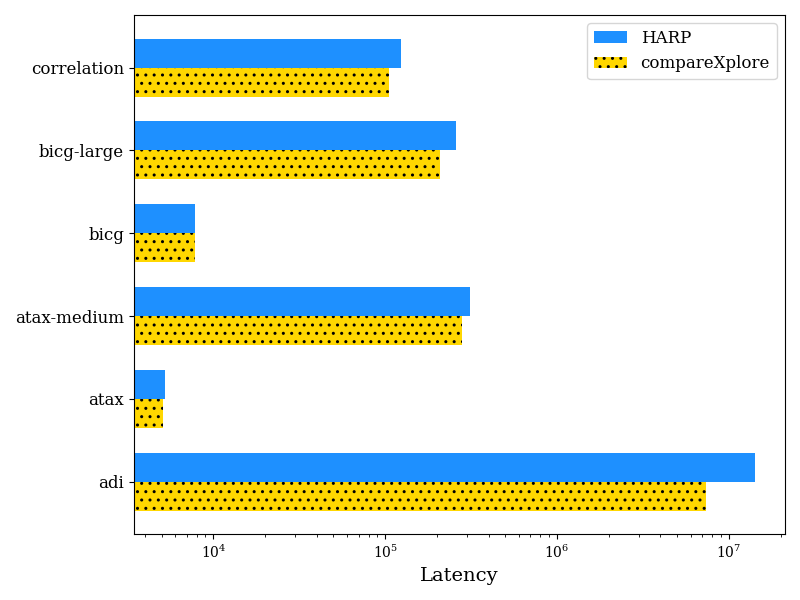}
    \vspace{-0.3cm}
    \caption{Latency in terms of cycle count ($\downarrow$) of the final designs selected by the DSE stage. The figure is on the \textbf{log-scale}.}
    \label{fig:latency}
\end{figure}

Figure~\ref{fig:latency} shows the design space exploration results using our proposed approach compared to the vanilla HARP model. The results demonstrate that our proposed approach consistently outperforms the vanilla HARP model across all kernels in terms of latency reduction. On average, \pmodel achieves a 16.11\% reduction in latency compared to \harp. The improvement is particularly significant for the ``adi'' kernel, where \pmodel reduces the latency by nearly 50\%. These results highlight the effectiveness of our approach in identifying high-quality designs that lead to improved hardware performance.

\vspace{-0.5cm}

\subsection{Parameter Sensitivity Study}

\begin{figure}[h]
    \centering
    \includegraphics[width=0.95\linewidth]{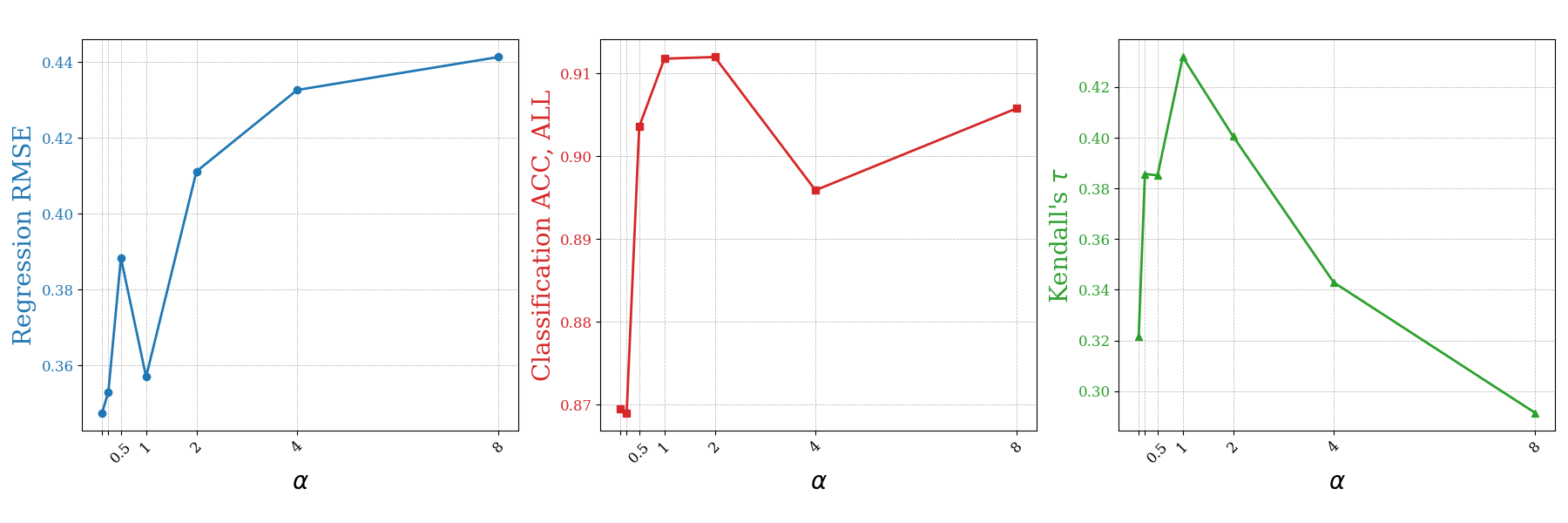}
    \vspace{-0.5cm}
    \caption{As $\alpha$ increases, the model places more emphasis on the pairwise loss compared to the pointwise loss. $\alpha$ varies in $\{0.125, 0.25, 0.5, 1, 2, 4, 8\}$.}
    \label{fig:alpha_study}
\end{figure}

Figure~\ref{fig:alpha_study} shows the effect of $\alpha$ on the \pmodel's performance. As $\alpha$ increases, the regression RMSE worsens, while the classification accuracy peaks around $\alpha=1$. Kendall's $\tau$ ranking metric reaches its highest value at $\alpha=1$ and then declines. These trends suggest that excessive emphasis on pairwise comparisons may not necessarily improve overall performance. In contrast, moderate $\alpha$ values effectively balance pointwise and pairwise losses, optimizing both tasks effectively.

\subsection{Time Breakdown Analysis}
The average time breakdown analysis presented in Table~\ref{tab:avg_time_breakdown} highlights the efficiency of our two-stage DSE process. On average, Stage 1 accounts for approximately 87.06\% of the total computation time, while Stage 2 contributes only 12.94\%. This demonstrates that the pairwise comparison phase (Stage 2) introduces minimal additional overhead, ensuring that the overall computational efficiency is maintained. The relatively small proportion of time spent in Stage 2 indicates that our approach is practical and scalable for large-scale design space exploration tasks, making it suitable for optimizing HLS designs.

\begin{table}[h]
\centering
\caption{Average Time Breakdown of Stage 1 and Stage 2}
\label{tab:avg_time_breakdown}
\begin{tabular}{lcc}
\toprule
Stage & Average Time (\%) \\
\midrule
Stage 1 & 87.06 \\
Stage 2 & 12.94 \\
\bottomrule
\end{tabular}
\end{table}

% These results demonstrate the effectiveness of our proposed approach in identifying high-quality designs through pairwise comparisons and ranking. The proposed \pmodel is able to capture critical pragma-related differences, leading to improved offline evaluation results on and online DSE results with a relatively small amount of running time overhead.

%% file: sec-conc.tex
\section{Conclusion and Future Work}
\label{sec-conc}

In this paper, we presented \pmodel, a novel approach for HLS design space exploration that addresses the challenges of modeling complex design performance relationships. By incorporating a hybrid loss function, a Node Difference Attention module, and a two-stage DSE approach, \pmodel demonstrates significant improvements in both pairwise comparison accuracy and ranking metrics. Our results show that explicitly learning to compare designs, with a focus on pragma-induced variations, leads to the discovery of higher quality HLS-generated designs.

Although \pmodel does not achieve the lowest regression error compared to \harp, our results show that explicitly learning to compare designs leads to the discovery of higher quality HLS-generated designs. In addition, in practice, it is worth considering a separate model such as \harp for stage 1 of the DSE process specializing in accurate pointwise prediction.

The success of \pmodel in HLS DSE highlights the broader potential of learning-to-rank methods in the hardware optimization domain. In future work, we believe that this paradigm can be further explored and extended. For example, the ability to rank and select designs within a large language model (LLM) framework could lead to tighter integration of language models and hardware design enabling a more intuitive and automated design process, and achieving better performance in both regression and classification due to the higher expressive power of LLMs in capturing complex design relationships and patterns.